\documentclass{article}

\usepackage[final]{tackling_climate_workshop_style}

\usepackage[utf8]{inputenc} 
\usepackage[T1]{fontenc}    
\usepackage{hyperref}       
\usepackage{url}            
\usepackage{booktabs}       
\usepackage{amsfonts}       
\usepackage{nicefrac}       
\usepackage{microtype}      
\usepackage[pdftex]{graphicx}
\usepackage{subfig}
\usepackage{floatrow}
\newfloatcommand{capbtabbox}{table}[][\FBwidth]

\title{Personalizing Sustainable Agriculture with Causal Machine Learning}

\author{%
  Georgios Giannarakis \\
  BEYOND Centre, IAASARS\\
  National Observatory of Athens\\
  \texttt{giannarakis@noa.gr} \\
  \And
  Vasileios Sitokonstantinou \\
  BEYOND Centre, IAASARS \\
  National Observatory of Athens \\
  \texttt{vsito@noa.gr} \\
  \AND
  Roxanne Suzette Lorilla \\
  BEYOND Centre, IAASARS \\
  National Observatory of Athens \\
  \texttt{rslorilla@noa.gr} \\
  \And
  Charalampos Kontoes \\
  BEYOND Centre, IAASARS \\
  National Observatory of Athens \\
  \texttt{kontoes@noa.gr}
}

\begin{document}

\maketitle

\begin{abstract}
To fight climate change and accommodate the increasing population, global crop production has to be strengthened. To achieve the ``sustainable intensification'' of agriculture, transforming it from carbon emitter to carbon sink is a priority, and understanding the environmental impact of agricultural management practices is a fundamental prerequisite to that. At the same time, the global agricultural landscape is deeply heterogeneous, with differences in climate, soil, and land use inducing variations in how agricultural systems respond to farmer actions. The ``personalization'' of sustainable agriculture with the provision of locally adapted management advice is thus a necessary condition for the efficient uplift of green metrics, and an integral development in imminent policies. Here, we formulate personalized sustainable agriculture as a Conditional Average Treatment Effect estimation task and use Causal Machine Learning for tackling it. Leveraging climate data, land use information and employing Double Machine Learning, we estimate the heterogeneous effect of sustainable practices on the field-level Soil Organic Carbon content in Lithuania. We thus provide a data-driven perspective for targeting sustainable practices and effectively expanding the global carbon sink.

\end{abstract}

\section{Introduction}

Climate change poses a multidimensional, complex challenge to humankind. In this context, agriculture faces a unique situation. On the one hand, production must keep rising up to meet ever-increasing demands [1], a task that is significantly complicated by inflation, crises, and extreme, unpredictable events [2,3,4]. On the other hand, it must do so in a sustainable way. Owing to decades of scientific research, the consequences of environmentally oblivious agricultural practices are by now well understood [5,6]. A unique opportunity also surfaced; since agricultural land comprises 38\% of the global land surface [7], agriculture offers a clear path towards mitigating climate change, by maximizing the carbon that crops naturally sequester, thus creating a substantial carbon sink [8].

As a result, insights on the impacts of sustainable agricultural practices, especially with regards to their effect on Soil Organic Carbon (SOC) are invaluable, not only for sustainability purposes, but also for the reliability, fairness, and transparency of modern agricultural carbon markets. At the same time, the global agricultural landscape contains significant diversity in terms of climate, soil, and land use. Such diversity of agricultural ecosystems frequently results in heterogeneous responses to interventions carried out by farmers; as an example, the effect of fertilizers on soil organic matter may differ depending on soil pH [9]. This reality is now recognized and reflected on high-stakes policy making. In an effort to ensure a sustainable future, the latest iteration of the Common Agricultural Policy (CAP) of the European Union explicitly provides greater flexibility for adapting its measures to local conditions [10]. 
From the perspective of artificial intelligence, such endeavor entails a personalization problem, where the end goal is to provide the farmer with targeted advice, depending on the particular farming context, on best practices to use in order to uplift sustainable metrics, such as SOC, and e.g., reimburse them accordingly. In this context, causal Machine Learning (ML) can efficiently leverage big Earth Observation (EO) data and other agricultural information in order to arrive at quantitative insights on the impacts of sustainable agriculture. Due to the large geographical coverage attained by EO, such methods can estimate effect heterogeneity even at the field-level, hence providing personalized and truly actionable advice towards sustainability goals. In this paper, we apply causal ML to understand the heterogeneous response of SOC content to a sustainable agriculture program and showcase a methodology for its personalization.

\section{Related Work}

Over the past decade, ML methods have made significant contributions in a variety of important tasks, including yield prediction, or pest, weed, and disease detection [11]. While such work has been fundamental in advancing precision agriculture, it is primarily based on predictive, correlation-based models, and is hence inadequate for assessing causal relations. Recently, works that aim to explicitly model cause and effect have been published. Enabled by large-scale EO data, among other sources, researchers have used new causal ML methods to infer the impact of conservation tillage on yield [12], or assess the suitability of agricultural land for applying crop rotation and diversification [13]. Causal forests [14] have been employed [15] for evaluating the environmental effectiveness of various agri-environment schemes [16]. However, results were obtained either on a lower spatial resolution, hindering the field-level personalization of advice [12, 13], or for a single year only [15], impeding the capture of slow, long-term environmental dynamics, such as those characterising soil [17]. Here, we utilize field-level SOC predictions and other geospatial data to propose a multi-year, fully personalized framework for assessing the effect of sustainable practices on SOC content and guide decision-making. We report preliminary results for a case study in Lithuania.

\section{Data and Methods}

\textbf{Dataset.} We use data from the Lithuanian Land Parcel Identification System (LPIS) that contains national-scale information on the crop type that was cultivated on each geo-referenced field as declared by the farmers in their CAP subsidy applications. Data also contain a binary variable that indicates if a field was enrolled in an eco-friendly program and employed sustainable agricultural management practices for that year. Data were collected annually, spanning a 5-year period from 2017 to 2021. In order to control for climate-induced confounding we incorporate 9 climate variables retrieved from the ERA5 database, namely air temperature, soil temperature and moisture, northward and eastward wind speed, evapotranspiration, leaf area index, runoff and precipitation [18]. 
To complete the data, we include field-level predictions of SOC content [19, 20], on which we want to examine the impact of sustainable practices. Due to the dependence of the SOC content prediction method on the absence of cloud cover, we currently only use a single SOC content value for the entire 5-year period. While increasing the temporal resolution of SOC content would provide more granular insights, we note that due to the slowly changing soil properties [17], the single value we have may potentially be decently representative of the reference period and subsets of it.

\textbf{Methods.} We model the problem of assessing the causal effect of eco-friendly practices on SOC content in different farming contexts as a Conditional Average Treatment Effect (CATE) estimation task. Using the Potential Outcomes framework, let $Y(T)$ denote the value (potential outcome) of a random variable $Y$ if we were to treat a unit with a binary treatment $T \in \{0, 1\}$. Given a vector $X$ describing the units, the CATE is the function $\theta(x) = \mathbb{E}[Y(1) - Y(0) | X = x]$. In our case, $T$ is the binary variable that indicates whether a field was enrolled in the eco-friendly program, and $Y(1), Y(0)$ are, respectively, the SOC content the field would have had if it was enrolled in the eco-friendly program ($T=1$) or not ($T=0$). The feature vector $X$ captures important field characteristics that drive effect heterogeneity. Here, to segment the analysis per crop type, $X$ consists of 3 binary variables, corresponding to the 3 most prominent crops found in the dataset during the reference period (Table \ref{tab:results}). We then use Double Machine Learning (DML) [21] to learn $\theta(x)$ from the dataset, and estimate the effect of sustainable practices on SOC content as a function of $x$.

\section{Results}

In a preliminary analysis, we designated as ``treated'' the fields that employed sustainable practices for the years 2021 and 2020, and as ``control'' the rest. We thus temporally aggregated all ERA5 climate variables over these two years, ending up with a single (average) value for each variable per field. We applied a standard scaler to all numerical features, and used a Random Forest Regressor and Classifier for the internal predictive DML tasks. To retain full inference on causal effect estimates, we used the Linear DML version, as implemented in [22].

\begin{figure}[!h]
\begin{floatrow}
\ffigbox{%
\hspace{-0cm}
  \includegraphics[scale=0.174]{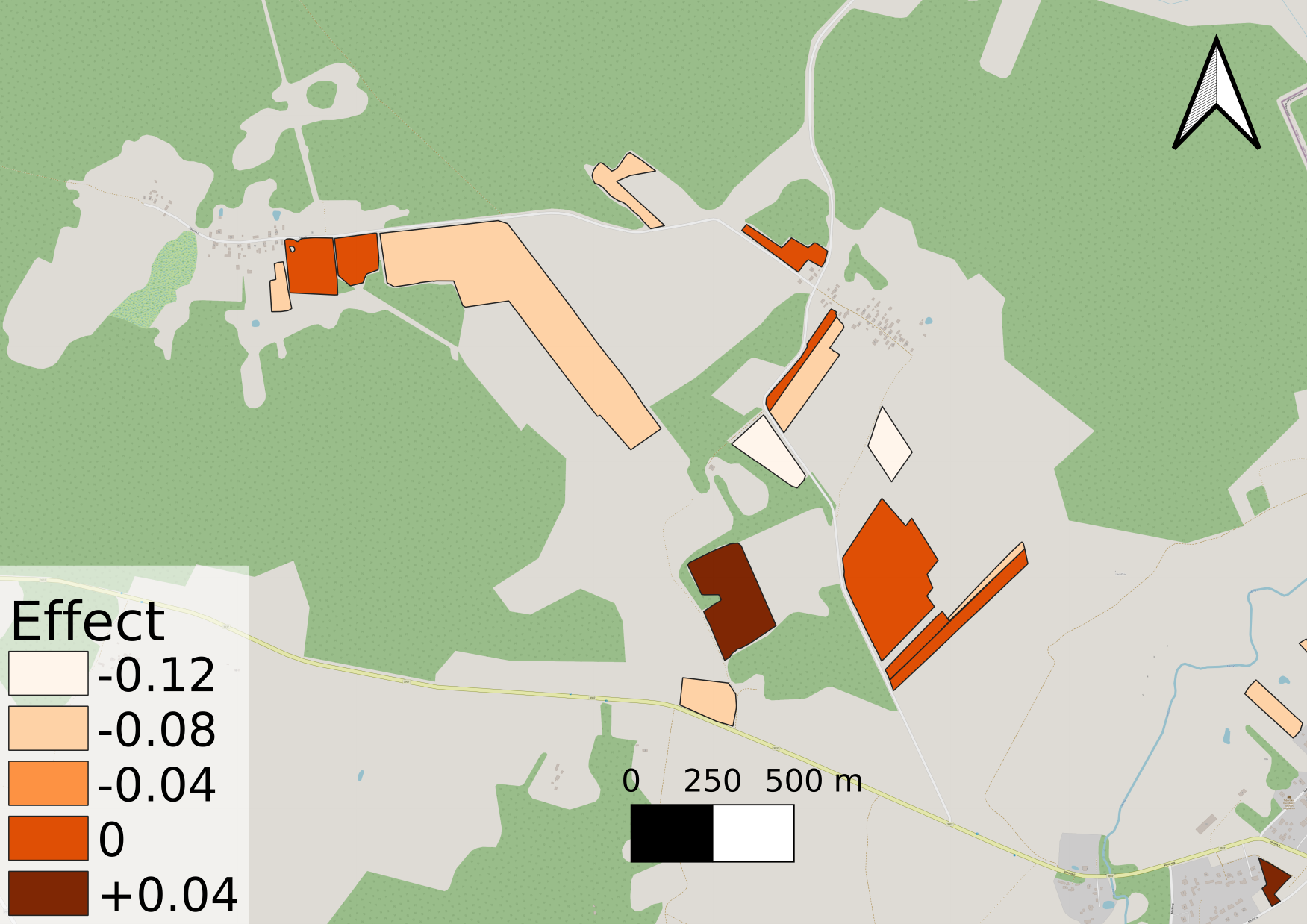}
}{%
\hspace{-0cm}
  \caption{Color coded CATE point estimates for selected fields. Dark fields have larger (non-significant) estimates and are expected to better accommodate sustainable practices in terms of their impact on SOC content.}%
  \label{fig:results}
}
\capbtabbox{%
    \begin{tabular}{lccc}
        \toprule
        \multicolumn{4}{c}{Conditional Average Treatment Effects}     \\
        \toprule
        Crops & PP & SP & WW   \\
        \midrule
        Point Estimate   & 0.06  & -0.08 & -0.09 \\
        Standard Error   & 0.14  & 0.17  & 0.28  \\
        Z-score      & 0.41  & -0.44 & -0.32  \\
        P-value          & 0.69  & 0.66  & 0.75   \\
        95\% CI (Lower)  & -0.22 & -0.42 & -0.63  \\
        95\% CI (Higher) & 0.34  & 0.26  & 0.45   \\
        \bottomrule
    \end{tabular}
}{%
  \caption{CATE point estimates and associated uncertainty. PP = Perennial Pastures, SP = Simple Pastures, WW = Winter Wheat. SOC content is reported as a percentage of topsoil (0-10 cm).}%
  \label{tab:results}
}
\end{floatrow}
\end{figure}

The most prominent crops found in the dataset consisted of fields with perennial pastures (PP) in both 2021 and 2020, followed by simple pastures (SP) and winter wheat (WW), also in both years. Table \ref{tab:results} contains the corresponding CATE estimates, quantifying the causal effect of sustainable practices on SOC content per crop, and Figure \ref{fig:results} visualizes them for a few indicative fields. None of the effects are currently statistically significant at the 95\% confidence level, a sensible result given the short time window the treatment refers to, its discrepancy with the 5-year period the SOC content values were derived from, and the slow rate of change of soil properties. As we incorporate more years and data in the analysis we will be able to explore interesting scientific questions, such as assessing whether the eco-friendly program was most efficient in perennial pastures and corroborate the growing evidence on the environmental benefits of perennialization [23].

\section{Conclusions and Future Work}

Our work can assist policy makers in efficiently implementing sustainable agricultural management and simultaneously enable the smart allocation of farmer subsidies, a task that has proven to be challenging [24]. Climate, soil, and other land use information may also be included in the data in order to adapt the estimates on any local condition experts deem important. In that way, the idea presented in this paper can develop into a useful tool for sustainable agriculture, in line with modern policy needs and encompassing the entire agricultural chain from the policy maker to the farmer.
To proceed, the temporal range of the treatment should be extended to align with the 5-year SOC content period, and domain knowledge should be incorporated in the model through structural approaches [25]. The inclusion of more data on field-level agricultural practices is also important for reducing the bias of estimates. Finally, given the general absence of ground truth, estimates should be tested for robustness by employing modern sensitivity analyses [26].

\begin{ack}
This work has been supported by the EIFFEL project (EU Horizon 2020 - GA No. 101003518) and the MICROSERVICES project (2019-2020 BiodivERsA joint call, under the BiodivClim ERA-Net COFUND programme, and with the GSRI, Greece - No. T12ERA5-00075).
\end{ack}

\section*{References}


\small

[1] Tilman, D., Balzer, C., Hill, J., \& Befort, B. L. (2011). Global food demand and the sustainable intensification of agriculture. Proceedings of the national academy of sciences, 108(50), 20260-20264.

[2] Sivakumar, M. V., Motha, R. P., \& Das, H. P. (Eds.). (2005). Natural disasters and extreme events in agriculture: impacts and mitigation. Berlin, Heidelberg: Springer Berlin Heidelberg.

[3] Gray, R. S. (2020). Agriculture, transportation, and the COVID‐19 crisis. Canadian Journal of Agricultural Economics/Revue canadienne d'agroeconomie, 68(2), 239-243.

[4] Brunelle, T., Dumas, P., Souty, F., Dorin, B., \& Nadaud, F. (2015). Evaluating the impact of rising fertilizer prices on crop yields. Agricultural economics, 46(5), 653-666.

[5] Tsiafouli, M. A., Thébault, E., Sgardelis, ... \& Hedlund, K. (2015). Intensive agriculture reduces soil biodiversity across Europe. Global change biology, 21(2), 973-985.

[6] Robertson, G. P., Paul, E. A., \& Harwood, R. R. (2000). Greenhouse gases in intensive agriculture: contributions of individual gases to the radiative forcing of the atmosphere. Science, 289(5486), 1922-1925.

[7] Food and Agriculture Organization of the United Nations (FAO), Land use in agriculture by the numbers, URL https://www.fao.org/sustainability/news/detail/en/c/1274219/

[8] Toensmeier, E. (2016). The carbon farming solution: A global toolkit of perennial crops and regenerative agriculture practices for climate change mitigation and food security. Chelsea Green Publishing.

[9] Heinze, S., Raupp, J., \& Joergensen, R. G. (2010). Effects of fertilizer and spatial heterogeneity in soil pH on microbial biomass indices in a long-term field trial of organic agriculture. Plant and Soil, 328(1), 203-215.

[10] European Commission, The new common agricultural policy: 2023-27, URL https://agriculture.ec.europa.eu/common-agricultural-policy/cap-overview/new-cap-2023-27\_en


[11] Liakos, K. G., Busato, P., Moshou, D., Pearson, S., \& Bochtis, D. (2018). Machine learning in agriculture: A review. Sensors, 18(8), 2674.

[12] Deines, J. M., Wang, S., \& Lobell, D. B. (2019). Satellites reveal a small positive yield effect from conservation tillage across the US Corn Belt. Environmental Research Letters, 14(12), 124038.

[13] Giannarakis, G., Sitokonstantinou, V., Lorilla, R. S., \& Kontoes, C. (2022). Towards assessing agricultural land suitability with causal machine learning. In Proceedings of the IEEE/CVF Conference on Computer Vision and Pattern Recognition (pp. 1442-1452).

[14] Wager, S., \& Athey, S. (2018). Estimation and inference of heterogeneous treatment effects using random forests. Journal of the American Statistical Association, 113(523), 1228-1242.

[15] Stetter, C., Mennig, P., \& Sauer, J. (2022). Using Machine Learning to Identify Heterogeneous Impacts of Agri-Environment Schemes in the EU: A Case Study. European Review of Agricultural Economics.

[16] Batáry, P., Dicks, L. V., Kleijn, D., \& Sutherland, W. J. (2015). The role of agri‐environment schemes in conservation and environmental management. Conservation Biology, 29(4), 1006-1016.

[17] Smith, P. (2004). How long before a change in soil organic carbon can be detected?. Global Change Biology, 10(11), 1878-1883.

[18] Hersbach, H., Bell, B., Berrisford, P., ... \& Thépaut, J. N. (2020). The ERA5 global reanalysis. Quarterly Journal of the Royal Meteorological Society, 146(730), 1999-2049.

[19] Tziolas, N., Tsakiridis, N., Ogen, Y., Kalopesa, E., Ben-Dor, E., Theocharis, J., \& Zalidis, G. (2020). An integrated methodology using open soil spectral libraries and Earth Observation data for soil organic carbon estimations in support of soil-related SDGs. Remote Sensing of Environment, 244, 111793.

[20] Karyotis, K., Samarinas, N., Tsakiridis, N., Tziolas, N., Kokkas, S., Tsiakos, V., Bliziotis, D., Kalopesa, E., \& Zalidis, G. (2022). Synergy between Spaceborne optical Imagery and in-situ Soil Spectroscopy for Organic Carbon estimations over exposed lands. 26th MARS Conference.

[21] Chernozhukov, V., Chetverikov, D., Demirer, M., Duflo, E., Hansen, C., Newey, W., \& Robins, J. (2018). Double/debiased machine learning for treatment and structural parameters.

[22] Battocchi, K., Dillon, E., Hei, M., Lewis, G., Oka, P., Oprescu, M., \& Syrgkanis, V. (2019) EconML: A Python Package for ML-Based Heterogeneous Treatment Effects Estimation.

[23] Mosier, S., Cordova, S., \& Robertson, G. P. (2021). Restoring soil fertility on degraded lands to meet food, fuel, and climate security needs via perennialization. Frontiers in Sustainable Food Systems, 356.

[24] European Court of Auditors. (2017). Greening: a more complex income support scheme, not yet environmentally effective. Special Report, (21).

[25] Pearl, J. (2009). Causality. Cambridge university press.

[26] Cinelli, C., \& Hazlett, C. (2020). Making sense of sensitivity: Extending omitted variable bias. Journal of the Royal Statistical Society: Series B (Statistical Methodology), 82(1), 39-67.




\end{document}